\documentclass[10pt,twocolumn,letterpaper]{article}

\pdfoutput=1

\usepackage{wacv}
\usepackage{times}
\usepackage{epsfig}
\usepackage{graphicx}
\usepackage{amsmath}
\usepackage{amssymb}
\usepackage{caption}
\usepackage{booktabs}
\usepackage{bm}
\usepackage{xcolor}

\wacvfinalcopy 

\def\wacvPaperID{****} 

\ifwacvfinal\fi
\begin{document}

\title{Treelogy: A Novel Tree Classifier Utilizing Deep and Hand-crafted Representations}

\author{\.{I}lke \c{C}u\u{g}u, Eren \c{S}ener, \c{C}a\u{g}r{\i} Erciyes, Burak Balc{\i}, Emre Ak{\i}n, It{\i}r \"{O}nal, Ahmet O\u{g}uz Aky\"{u}z\\
Department of Computer Engineering, Middle East Technical University\\
{\tt\small \{cugu.ilke, sener.eren\}@metu.edu.tr; \{e1881176, e1881044, e1880897, itir, akyuz\}@ceng.metu.edu.tr}
}

\maketitle
\ifwacvfinal\thispagestyle{empty}\fi

\begin{abstract}

   We propose a novel tree classification system called \emph{Treelogy}, that fuses deep representations with hand-crafted features obtained from leaf images to perform leaf-based plant classification. Key to this system are segmentation of the leaf from an untextured background, using convolutional neural networks (CNNs) for learning deep representations, extracting hand-crafted features with a number of image processing techniques, training a linear SVM with feature vectors, merging SVM and CNN results, and identifying the species from a dataset of 57 trees. Our classification results show that fusion of deep representations with hand-crafted features leads to the highest accuracy. The proposed algorithm is embedded in a smart-phone application, which is publicly available. Furthermore, our novel dataset comprised of 5408 leaf images is also made public for use of other researchers.
\end{abstract}

\section{Introduction}

The classification work on plants is as old as humankind. Ancient civilizations such as Assyrians, Egyptians, Chinese and Indians were interested in identification of plant species, especially for medical purposes. When Theophrastus of Eresus completed his work \textit{Historia Plantarum}, he opened the doors leading to modern plant taxonomy. Through ages, scientific discoveries have shaped the means of plant classification, and the works on creation of machine intelligence revealed a new question: is it possible to develop an effective computer-based plant classifier?

Earlier studies focused on finding the most appropriate features of leaf images using image processing. They trained classifiers with hand-crafted feature vectors for tree classification. However, the emerging success of convolutional neural networks (CNNs) in classification tasks caused a shift of interest towards methods of tree identification. Pure CNN approaches are proved to be powerful against classifiers trained only with hand-crafted features. Furthermore, recent experiments have shown that combining CNNs with hand-crafted features generally improves performance with respect to both used in isolation~\cite{hall2015evaluation}.

\begin{figure}
\includegraphics[scale=0.3]{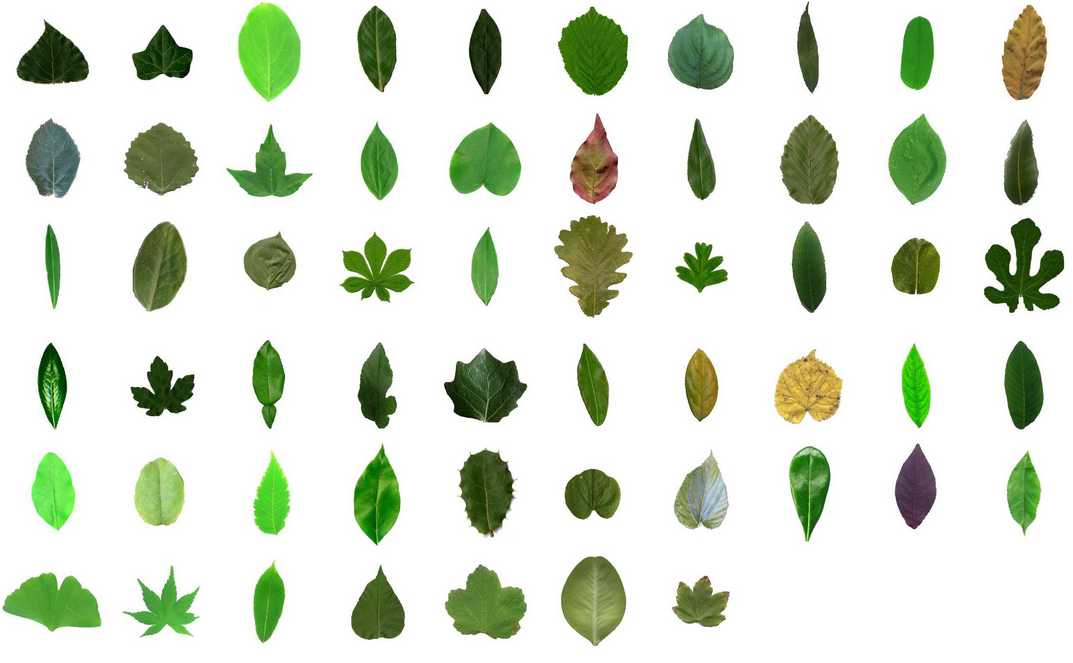}
\caption{Sample of the 57 tree species employed in this paper.}
\label{fig:57_leaf_species}
\end{figure}

\begin{figure*}
\centering
\includegraphics[scale=0.313]{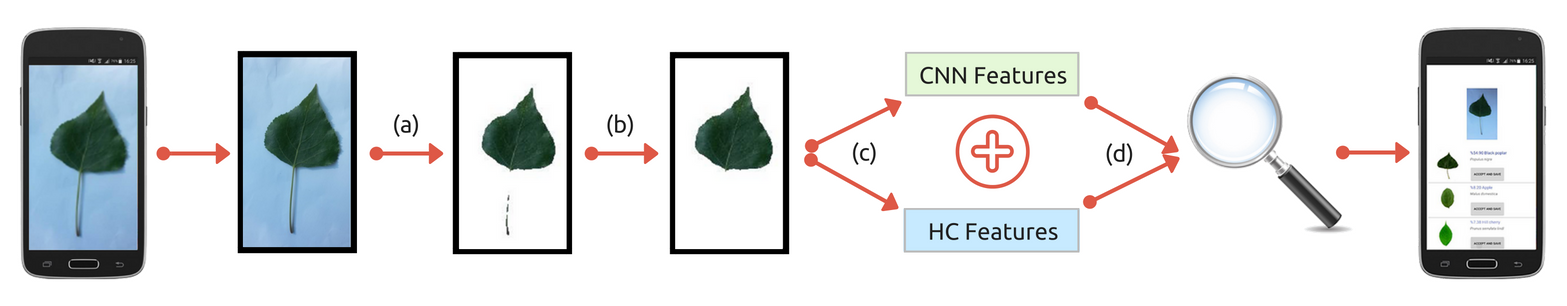}
\centering
\captionsetup{justification=centering}
\caption{Lifecycle of Treelogy using a \textit{populus nigra} leaf: (a) Background elimination (b) Stem removal (c) Extraction of hand-crafted and ConvNet features (d) Predictions from fused feature vectors}
\label{fig:lifecycle}

\end{figure*}

In this work, we describe a visual recognition system for leaf based tree identification. The system, called \emph{Treelogy}, helps users identify trees from an image of a given leaf. In Treelogy, images first pass through a pre-processing step in which background elimination and stem removal operations are applied. Then, hand-crafted features are extracted from pre-processed images using image processing techniques. Moreover, deep representations of images are learned using a CNN.  Finally, deep representations are combined with hand-crafted leaf features to classify $57$ tree species shown in Figure~\ref{fig:57_leaf_species}. The overall workflow of our approach is illustrated in Figure~\ref{fig:lifecycle}.

This paper presents three contributions. First, we propose an effective tree classification system which is capable of accurately discriminating between $57$ tree species, a number which is beyond the number of classes used in most of the earlier studies. Second, we share a leaf image dataset comprised of thousands of images which is put into public domain for facilitating research in this direction. The dataset consists of both new images taken by the authors and existing images gathered from various other public databases. This dataset is already preprocessed so that future algorithms may directly focus on classification, rather than starting with preprocessing tasks such as background elimination and stem removal. Finally, we implement our algorithm within an Android based smart-phone application, which can be readily downloaded and used by ecologists, amateur botanists, educators, children, and nature enthusiasts in general.

\section{Related Work}

Automatic tree identification from leaf images has been investigated for several decades. At the highest level, these studies can be categorized as those based on hand-crafted features and those that use neural networks (NNs).

Among the first category, features may be computed from 2D regions within the leaf image (region-based)~\cite{zhang2004review} or from the contour of the leaf (contour-based)~\cite{belongie2002shape}. Abbasi et al.~\cite{abbasi1997reliable} used Curvature Scale Space (CSS) technique to form their feature vectors and k-NN classifiers to classify chrysanthemum leaves. Kalyoncu and Toygar~\cite{kalyoncu2015geometric} used geometric features, Multi-Scale Distance Matrix, moment invariants and shape descriptors, then employed Linear Discriminant Classifier since it can work with different classes having different importance factor for features. Mouine et al.~\cite{mouine2013shape} gathered multiscale triangle representations of leaf images, and classified plants with k-NN classifiers. Another study for in-situ leaf recognition~\cite{olsen2015situ} extracts hand-crafted features, obtained by image processing methods, then use them to identify \textit{Lantana camara} with binary classification. Cao et al.~\cite{cao2016similarity} uses the curvature of the contour computed by overlaying circles of varying radii at various sampling points to compute a feature known as \emph{R-angle}. While some of these studies also yield promising results, they may fail to recognize distinctive properties of species with similar shapes due to the limited nature of hand-crafted features. 

There is an increasing tendency to use NNs in classification tasks for plant classification. Research methods are divided into three main groups. The first group of studies train NN classifiers using hand-crafted features. Chaki and Parekh~\cite{chaki2011plant} use image processing for feature extraction, then classify species with NN. Kadir et al.~\cite{kadir2013leaf} used Probabilistic Neural Network (PNN) to classify 32 tree species using shape, color, and texture features. The second group of studies employ pure CNNs for tree classification~\cite{reyes2015fine,lee2015deep}. Both studies preferred fine-tuning a pre-trained architecture to produce effective classifiers for their specific task. Among these, Reyes et al.~\cite{reyes2015fine} uses images of the entire plant, branches, fruits, stems, flowers and the leaves for classification. Lee et al.\cite{lee2015deep} classifies $44$ species using CNNs and finds that features learned through CNN outperform hand-crafted features. The authors furthermore find that cropping a region within the leaf image such that the outer edges of the leaf are invisible improves the quality of the CNN classifier. The authors also compare their work with the state-of-the-art, but the details of how exactly this comparison was performed are not elaborated. The third group of studies follow a hybrid approach where authors fuse both hand-crafted and CNN features to identify 32 species~\cite{hall2015evaluation}. Our work can be considered in the third group. The main difference between our work and that of Hall et al. \cite{hall2015evaluation} is that, they combine the separate probabilities of CNN and a classifier they called HCF-ScaleRobust via the sum rule, where we combine feature vectors of CNN and hand-crafted features obtained using image processing methods to produce a new feature vector to be used in LSVM training.   

Another aspect of plant recognition is background elimination. Since all studies above uses leaf images for identification, eliminating the background becomes an essential part of the process, and there are different approaches for this task based on whether the image has a textured or non-textured background. While~\cite{du2007leaf,fu2006combined,kadir2013leaf,beghin2010shape} segment leaves from simple, non-textured backgrounds using RGB and HSV thresholding, another study~\cite{olsen2015situ} focuses on textured backgrounds. Due to the nature of our dataset, we focus on segmentation of leaf from a non-textured background. 

It is worth noting that some studies developed mobile applications to serve their approaches through which users can benefit from the automated leaf classification algorithms. For example, LeafSnap~\cite{leafsnap_eccv2012} which is developed for iOS, is one of the first released applications of this kind. Subsequently, Pl@ntNet~\cite{goeau2013pl} mobile application, a more sophisticated one, has been released. A similar application MobileFlora~\cite{angelova2012development}, which is for flower recognition, is also relevant. We also built a mobile application to showcase our tree identification system. This application can be freely downloaded from Google Play Store\footnote{https://play.google.com/store/apps/details?id=com.payinekereg.treelogy}.

\section{Methodology}

\subsection{Background Elimination}
\label{ScBG}

This process aims to get eliminate non-leaf pixels of the image. To this end, we applied the k-means algorithm to cluster our images where the number of clusters was set as $k=2$ (leaf and non-leaf). As the distance function is based on color differences, we chose to work on the LAB color space rather than the highly correlated RGB space~\cite{Reinhard2008}. 

In this space, color information resides in the $AB$ plane with the lightness information stored along the $L$ axis. We empirically determined a threshold value in $A$ (red-green) channel to decide which bands to use in the Euclidean distance metric. When the average of ‘A’ band is higher than the threshold value, we employed the $AB$ plane for clustering, otherwise the $LA$ plane is used. The intention was to reduce the diverse illumination effect on the image.

For efficiency, the input image is scaled proportionally in such a way that the longer edge of the image becomes $256$ pixels. Then, k-means clustering is applied, where $k=2$, with three iterations using Euclidean distance. After that, the cluster corresponding to the leaf is identified.

\begin{figure}
\centering
\includegraphics[scale=0.2]{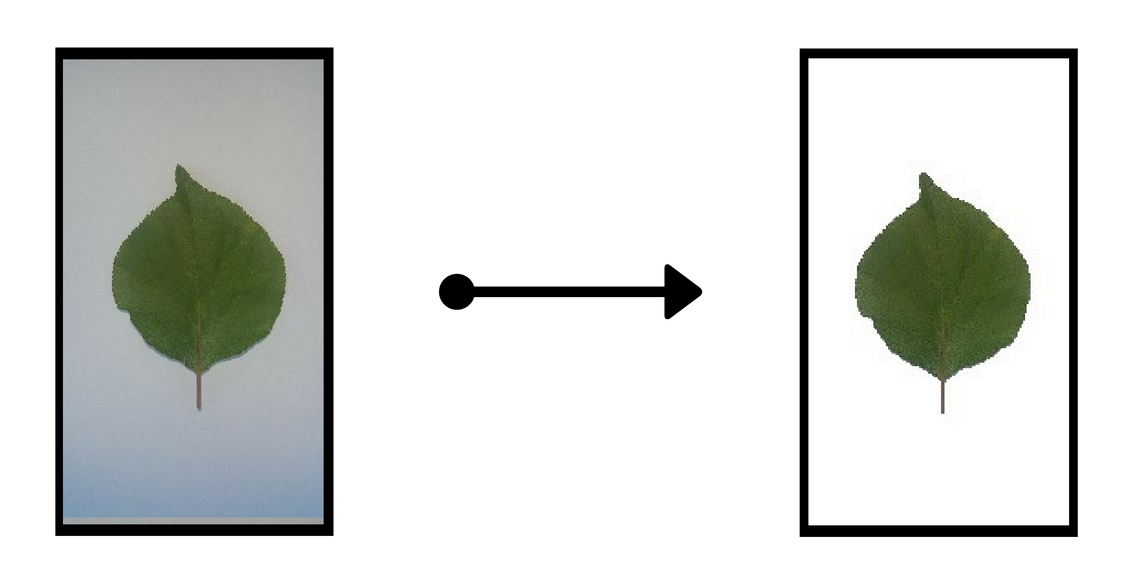}
\caption{Background elimination result of a \textit{prunus armeniaca} leaf.}
\label{fig:bg_elimination}
\end{figure}

Next, the original image is segmented out from non-leaf clusters. Since k-means returns noisy leaf clusters which have negative effects on feature extraction, the resulting image is converted into grayscale and a small amount of Gaussian smoothing is applied.  We decided to produce a binary image to use as a mask for segmentation. Therefore, inverse binary thresholding is applied with Otsu's binarization~\cite{otsu1975threshold} in order to get a binary image from the original one to be used as a alpha-mask. Finally, hole filling operation is applied on the obtained binary image containing the leaf cluster to bring back the wrongly clustered inner leaf pixels. The results of background elimination on a sample leaf image is shown in Figure~\ref{fig:bg_elimination}.

\begin{figure}
\centering
\includegraphics[scale=0.4]{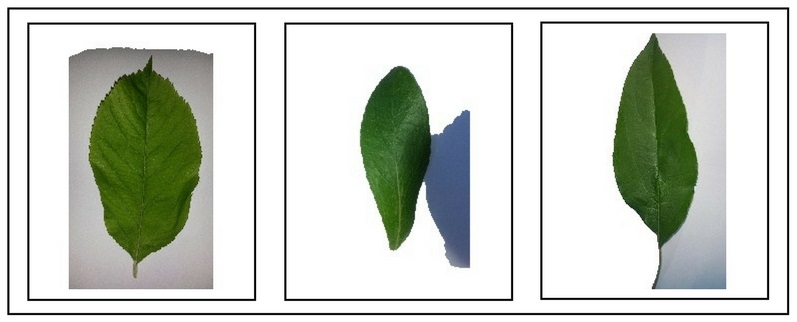}
\caption{Sample erroneous outputs of background elimination.}
\label{fig:bg_elimination_error}
\end{figure}

This method is tested with $310$ leaf images. Although the test images are taken using white papers as background, they contain shadows and heavy lightning effects, such that in some cases background appears to be gray, light blue, or yellowish. Table~\ref{tab:segmentation} shows the error rates of background elimination on different species. Outputs are considered erroneous when the background elimination method wipes out the entire image or leaves large pieces of the background untouched (Figure~\ref{fig:bg_elimination_error}). The overall accuracy is calculated as $85.5\%$.

\begin{table}
\caption{Error rates of background elimination for different species}
\label{tab:segmentation}
\begin{center}

\begin{tabular}{lc}
\toprule
Species & Error Rate \\
\midrule
Cercis siliquastrum			&	${0/3}$		\\
Hedera helix 				& 	${0/4}$ 	\\
Populus nigra				&	${0/4}$		\\
Magnolia grandiflora		&	${0/4}$		\\
Prunus avium				&	${1/4}$		\\
Aesculus hippocastanum 		& 	${2/4}$ 	\\
Nerium oleander				&	${0/5}$		\\
Corylus avellana			&	${1/7}$		\\
Quercus ilex				&	${0/9}$		\\
Tilia cordata				&	${3/42}$	\\
Prunus armeniaca			&	${0/45}$	\\
Salix babylonica			&	${4/61}$	\\
Malus domestica				&	${18/90}$	\\
Unknown 					& 	${16/28}$ 	\\
\midrule
Total						&	${45/310}$	\\
\bottomrule
\end{tabular}
\end{center}
\end{table}

\subsection{Stem Removal}
\label{ScSR}

\begin{figure}
\includegraphics[scale=0.2]{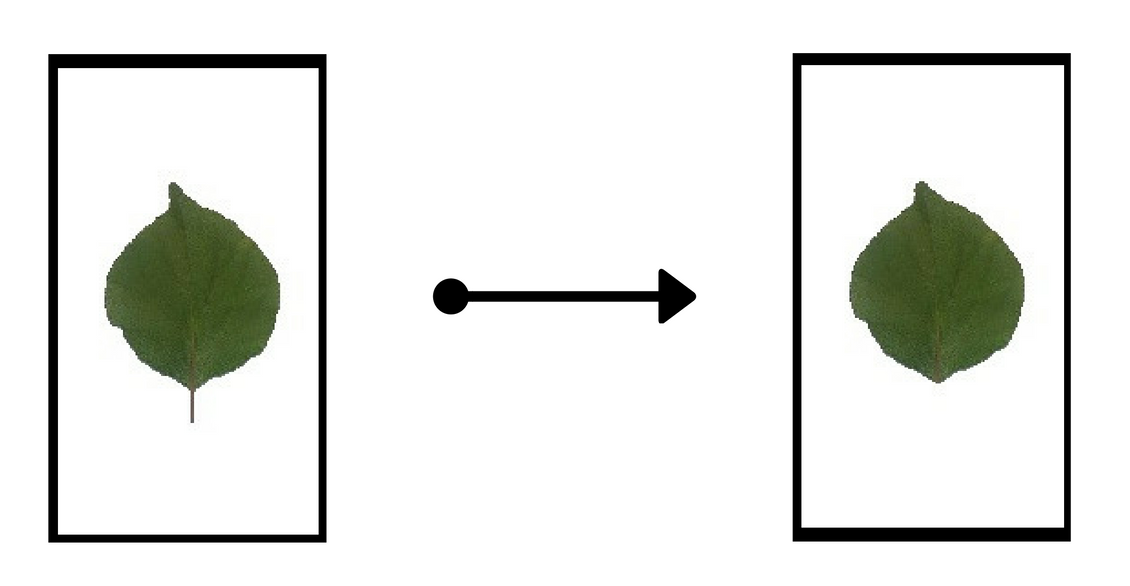}
\caption{Stem removal result of a \textit{prunus armeniaca} leaf.}
\label{fig:stem_removal}
\end{figure}

In order to remove the stem from the leaf, we used the opening operation which is an erosion followed by dilation~\cite{sonka2014image}. An ellipse shaped kernel with size parameter $(9, 9)$ is used as the instrument of the chosen morphological operation. We decided on the kernel size such that we can remove relatively thick stem pieces, while the deterioration of the leaf is still negligible. Notice that, we applied stem removal operation on the binary mask obtained after the operations described at background elimination step. A sample result of this process is illustrated in Figure~\ref{fig:stem_removal}.

\subsection{Feature Extraction}

\subsubsection{Extraction of Hand-crafted Features (HCFs)}

We used several image processing techniques to acquire a total of $56$ features from a leaf image. These were shape, contour, color, and texture features.

Note that, for texture features (HCF 7,8,9 and 10), the texture characterization is based on Gebejes and Huertas's proposal~\cite{gebejes2013texture}, where it is computed through second order statistical measurements based on Grey-Level Co-occurrence Matrix (GLCM). The GLCM contains information about the frequency of occurrences of two neighboring pixel combinations. Thus, two parameters, orientation and displacement, should be decided to set the correct neighboring between pixels in an image. We use horizontal orientation (0 degree) and a displacement value of $1$ in that direction. 

Our handcrafted features are the following:
\begin{enumerate}
\item \textbf{Ovality}: (Scalar) Ellipticity, result of semi-major axis divided by semi-minor axis.
\item \textbf{Area per Length}: (Scalar) Total inside area of the contour divided by total contour distance.
\item \textbf{Convexity}: (4x1 vector) The tendency of leaf contour to have convex structure. It gives defects on the contour between enveloped convex polygon.
\item \textbf{Solidity}: (Scalar) Total inside area of the contour divided by total area of the bounding convex hull area. It gives information about zigzag structure of the contour. 
\item \textbf{Equi-diameter}: (Scalar) The diameter of the circle, whose area is the same as the contour area.
\item \textbf{Extent}: (Scalar) The ratio of the contour area to the bounding rectangle's area.
\item \textbf{Correlation}: (Scalar) The measure of gray-level dependence between the pixels~\cite{albregtsen2008statistical}.
\item \textbf{Contrast}: (Scalar) The measure of local intensity variation on GLCM of the image~\cite{gebejes2013texture}.
\item \textbf{Entropy}: (Scalar) The measure of spatial disorder on GLCM of the image~\cite{gebejes2013texture}.
\item \textbf{Energy}: (Scalar) The measure of uniformity of texture on GLCM of the image~\cite{gebejes2013texture}.
\item \textbf{Standard Deviation}: (Scalar) The standard deviation of the normalized gray-level image.
\item \textbf{Mean}: (Scalar) The mean of normalized gray-level image.
\item \textbf{Corner Number}: (Scalar) The number of points that have a higher sharpness than a threshold value. It is computed using Harris corner detection method~\cite{harris1988combined}. 
\item \textbf{HU Moments}: (4x1 vector) The first four Hu~\cite{hu1962visual} invariant moments, which are invariant to rotation, translation and scaling. Normalized gray-level image is used for computation.
\item \textbf{Center Contour Distance}: (36x1 vector) Features that are gathered from Chaki and Parekh's~\cite{chaki2011plant} Centroid-Radii model. They define the shape of the leaf as normalized Euclidian distances from the contour pixels of the leaf to the contour center in counterclockwise order. In order to make the vector length constant, $36$ contour points are selected by dividing the contour to $36$ equal intervals starting from the contour pixel that has the $0$ degree angle with the contour center.
\end{enumerate}

\begin{figure}
\includegraphics[scale=0.8]{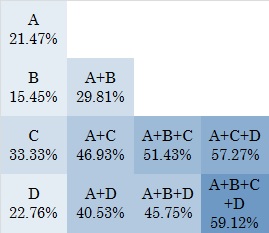}
\centering
\caption{Classification Performances of Employing Different Combinations of Handcrafted Features on Treelogy Dataset.}
\label{fig:hcf_precision}
\end{figure}

We split these handcrafted features into four groups to measure their participation to classification. We trained LSVM for $11$ combination of the groups on our dataset. Based on the results, we decided to use all hand-crafted features defined above. Figure \ref{fig:hcf_precision} shows accuracy results of the trials, and the groups are explained below:
\begin{itemize}
\setlength\itemsep{0.1em}
\item Group A (36x1): HCF 15 
\item Group B (8x1): HCFs 3 and 14
\item Group C (6x1): HCFs 1, 2, 4, 5, 6 and 13
\item Group D (6x1): HCFs 7, 8, 9, 10, 11 and 12
\end{itemize}

\subsubsection{Feature Extraction Using CNNs}

\begin{figure}
\includegraphics[scale=0.4]{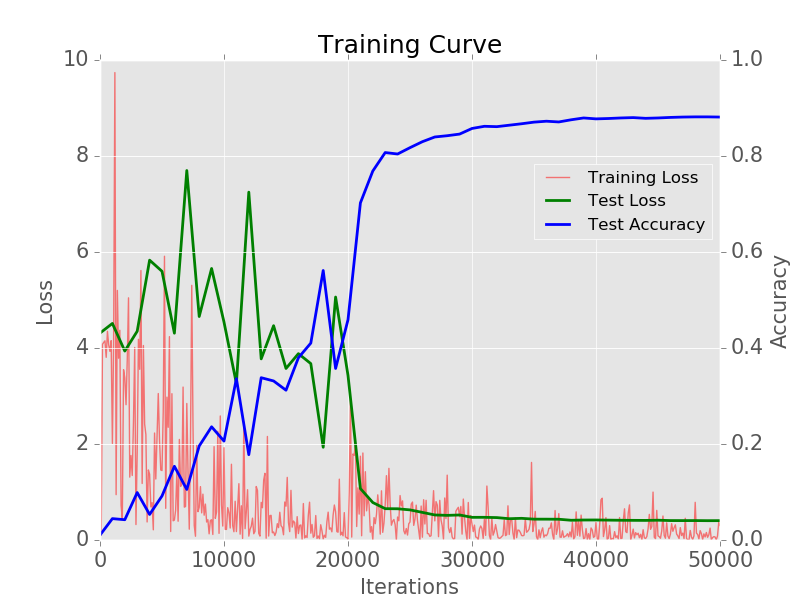}
\caption{Learning curve through $50000$ iterations.}
\label{fig:learning_curve}
\end{figure}

We used the Caffe Framework~\cite{jia2014caffe} pre-trained with Imagenet~\cite{ILSVRC15} as our CNN architecture. It is fine-tuned with our dataset to identify 57 tree species. BVLC-distributed (Berkeley Vision and Learning Center) CaffeNet model\footnote{https://github.com/BVLC/caffe/tree/master/models/bvlc\_reference\_caffenet} which is a slightly modified version of AlexNet~\cite{krizhevsky2012imagenet} is used to assign the initial weights. Stochastic Gradient Descent~\cite{bottou2012stochastic} is adopted as the optimization algorithm where the solver addresses the optimization of loss minimization computed for a mini-batch of N instances. Specifically, the average loss $L(W_t)$ is computed on the mini-batch where $f_w(X^{(i)})$ is the loss on data instance $X^{(i)}$ (Eq.\ref{eq:1}). Then, the negative gradient $\nabla L(W_t)$, the weight decay $\lambda$~\cite{krogh1992simple}, the learning rate $\alpha$ and the momentum $\mu$ are used to calculate $V_{t+1}$ (Eq.\ref{eq:2}) to update the current weight $W_t$ (Eq.\ref{eq:3}). In our case, the base learning rate ($\alpha_0$) is set as ${10^{-4}}$, and it is dropped by a factor of $10$ (Eq.\ref{eq:4}) after every $20000$ iterations. The dropout~\cite{hinton2012improving} is set as ${0.5}$, $\mu=0.9$, $\lambda=5*10^{-4}$ and the mini-batch size is $50$. The final Caffe model we used is constructed after $50000$ iterations (the learning curve of our fine-tuned model is shown in Figure \ref{fig:learning_curve}). 

\begin{figure*}
\centering
\includegraphics[scale=0.6]{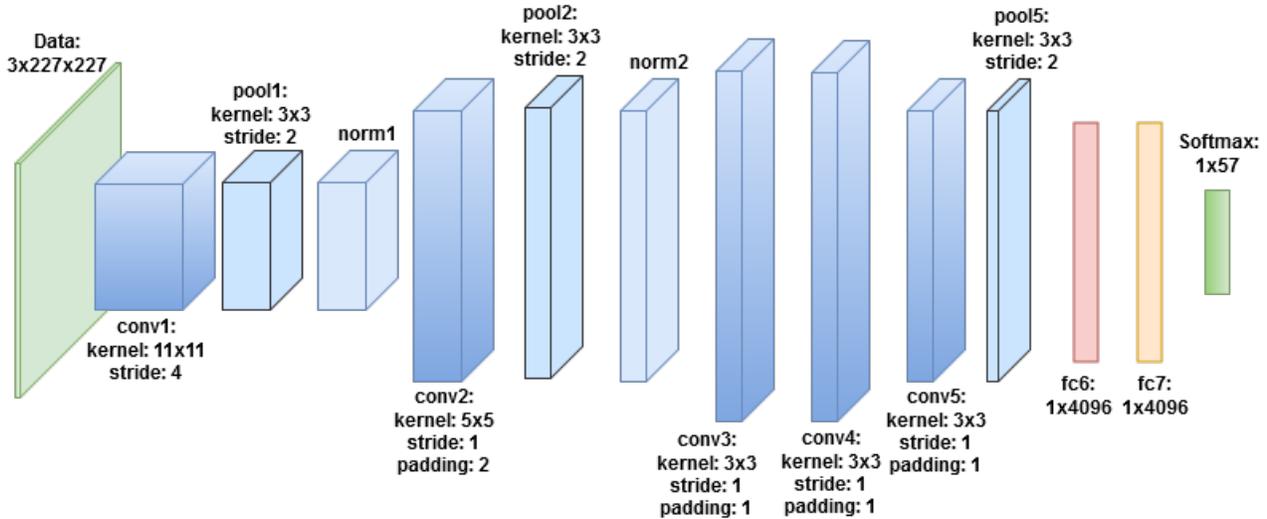}
\caption{Architecture of the CNN model for tree identification.}
\label{fig:architecture}
\end{figure*}

After the model is constructed, we used feature vectors of size $1\times4096$ obtained from the $fc6$ and $fc7$ layers of the architecture in our experiments  (see Figure \ref{fig:architecture}).

\begin{equation} \label{eq:1}
L(W_t)\gets \dfrac{1}{N} \sum_i^N f_w(X^{(i)})
\end{equation}
\begin{equation} \label{eq:2}
V_{t+1}\gets \mu V_t - \alpha (\nabla L(W_t) + \lambda W_t )
\end{equation}
\begin{equation} \label{eq:3}
W_{t+1}\gets W_t + V_{t+1}
\end{equation}
\begin{equation} \label{eq:4}
\alpha\gets \gamma^{\left \lceil{iter / step}\right \rceil} \alpha_0
\end{equation}

\section{Experiments and Results}

Naturally, the fundamental instrument of classification experiments is the dataset. The majority of leaf images in our dataset are taken from PlantNet~\cite{goeau2013pl}, Flavia~\cite{wu2007leaf} and LeafSnap~\cite{leafsnap_eccv2012} datasets, the rest are the product of our own photoshooting. In total, we gathered $4748$ training and $660$ testing images for $57$ tree species (Figure~\ref{fig:57_leaf_species}). To further expand our dataset, we rotated the images with three different angles (90$^{\circ}$, $180^{\circ}$, $270^{\circ}$) and enhanced our original dataset with the rotated images. Therefore, the total number of training and testing images were $18992$ and $2640$ respectively. All images were seperated from their backgrounds and stem-removed using the techniques described in Sections~\ref{ScBG} and~\ref{ScSR} with the exception of \textit{juglans regia} class which is used to introduce noise to the dataset. All leaf images were centered within the image region to minimize the impact of the leaf position within the original image.

In this study, we employed six different feature sets to classify trees from their leaf images. We trained an LSVM~\cite{fan2008liblinear} with hand-crafted features and the feature vectors obtained from the $fc6$ and $fc7$ layers as the first three approaches. Subsequently, we combined hand-crafted features with features of $fc6$ and $fc7$ layers separately and used them for LSVM trainings. In our final experiment, we use only the fine-tuned Caffe model's predictions.

Separation of training and testing images is a critical factor. The training set should be large in numbers and involve distinct images from various environments, sizes, ages, etc. while the test set should also have variations among leaf samples, so that it would be able to provide realistic accuracy estimations. We tried to find a balance between these two sets. Therefore, while we determined the test images, we used $5$ images for species with low variation of leaf images ($40$ - $60$ leaf images per species, photos taken in the same or two different environments), $10$ images for species with relatively more variation than the former ($70$ - $100$ leaf images per species, more photo sources than the former, but still not more than five different environments), and $15$ images for species with high variation ($>100$ leaf images per species, easy to find species with photos from wide range of environments). Then, we rotated the images with three different angles (90$^{\circ}$, $180^{\circ}$, $270^{\circ}$), and finalized our test set.

\begin{table}
\caption{Classification Performances of Employing Different Methods on Treelogy Dataset.}
\label{tab:classification}
\begin{center}

\begin{tabular}{lc}
\toprule
Method & Accuracy \\
\midrule
$\hbox{Hand-crafted}  \rightarrow  \hbox{LSVM}$                & 59.12\%  \\
$\hbox{fc6}  \rightarrow  \hbox{LSVM}$                        & 90.37\%  \\
$\hbox{fc7}  \rightarrow  \hbox{LSVM}$                        & 89.62\%  \\
$\hbox{\textbf{fc6 + Hand-crafted}} \rightarrow  \hbox{\textbf{LSVM}}$ & \textbf{90.5\%} \\
$\hbox{fc7 + Hand-crafted}  \rightarrow  \hbox{LSVM}$          & 89.58\%  \\
CNN                        								 	  & 84.62\%	 \\
\bottomrule
\end{tabular}
\end{center}
\end{table}

Finally, while combining CNN based feature vectors with hand-crafted features, we did not assign any weights on the features in order not to give an emphasis on a particular feature set. Therefore, the resulted vectors formed $4152 \times 1$ dimensional homogeneous feature vectors. 

The classification performances of utilizing six different feature sets can be found in Table~\ref{tab:classification}. Note that, this table represents top-1 accuracy values. It can be seen that the combination of hand-crafted features with deep representations lead to better top-1 accuracy compared to employing a single type of feature vector. We obtain the best top-1 accuracy ($\bm{90.5\%}$) among $57$ classes when we combine hand-crafted features with features obtained from $fc6$. At this point, it is important to note that CNN has $\bm{97.58\%}$ top-5 accuracy, which can be considered as superior over LSVM. Nonetheless, in the light of these results, we propose that combining CNN feature vectors with hand-crafted ones provides the most accurate tree classification model.  

\begin{table}
\caption{Classification Performances of Employing Different Methods on Flavia Dataset. (*Authors did not give the exact numbers in the related paper.)}
\label{tab:comparison}
\begin{center}

\begin{tabular}{ll}
\toprule
Method & Accuracy \\
\midrule
GLC~\cite{kalyoncu2015geometric}     									& *90+\%  \\
Flavia~\cite{wu2007leaf}     											& 90\%  \\
HCF-ScaleRobust~\cite{hall2015evaluation}                      			& 91.4\%  \\
HCF-ScaleRobust + ConvNet~\cite{hall2015evaluation}                     & 97.9\%  \\
CNN                        								 	 			& 98.44\%  \\
$\hbox{Hand-crafted}  \rightarrow  \hbox{LSVM}$          	 			& 92.18\%  \\
$\hbox{\textbf{fc6 + Hand-crafted}} \rightarrow  \hbox{\textbf{LSVM}}$ & \textbf{99.68\%} \\
\bottomrule
\end{tabular}
\end{center}
\end{table}

We also compare our results with recent works using Flavia dataset in Table~\ref{tab:comparison} to provide more insight on our model. The CNN model we used for Flavia dataset is built via fine-tuning our base CNN model. For fine-tuning, we initialized the weights using our fine-tuned base model, and the learning rate as ${10^{-4}}$. We dropped the learning rate by a factor of $10$ after every $1000$ iterations. We stopped fine-tuning after $3000$ iterations since the learning curve became nearly flat. Then, we merged this model's $fc6$ layer output with the hand-crafted features we introduced to finalize our proposed model for Flavia dataset. It can be seen that the proposed model has superior performance than the state-of-the-art solutions. Flavia dataset contains a small number of leaf images, $1907$ leaf images in total for $32$ species, with nearly identical intra-class and relatively distinguishable inter-class leaf images. Having such high classification results obtained from this dataset indicates that leaf classification task needs more challenging preprocessed datasets, as we proposed in this study\footnote{Our latest dataset can be found at http://treelogy.info/dataset}.

\section{Conclusions and Future Work}

In this paper, we demonstrate how to apply image processing and machine learning techniques to discriminate between $57$ different tree species with accuracy rates outperforming the state-of-the-art. We proposed that CNNs can be used to provide a large set of representative features for identification, and when combined with hand-crafted features, training LSVM outperforms pure CNN classification approach. We also introduce a preprocessed leaf image dataset which can further stimulate research in leaf-based tree identification. Our algorithms are implemented within a real-world smart phone application, and we invite the readers to try out this application.

Several future research directions present themselves. First, to improve the performance geographical location from the user's smart phone can be retrieved. This would allow ruling out impossible specifies given the user's current location. Second, a weight learning scheme can be incorporated to weigh the relative importance of CNN and hand-crafted features. While the LSVM naturally explores such relationships, given the higher number of CNN features, a global weighting parameter to balance between these two types of features may be beneficial. Furthermore, deep learning is a rapidly advancing field of research. Therefore, other types of CNN models may prove to be beneficial for leaf-based tree recognition.

As for more applied research directions, implementing such a system as an augmented reality application within a head-mounted display and providing real-time feedback to a nature explorer about the species within her field-of-view could be immensely useful. Likewise, one could deploy a quadcopter type drone to rapidly explore the flora present in a selected region of interest. Both problems would require more robust methods of background estimation due to heavy clutter in natural habitats.

{\small
\bibliographystyle{ieee}
\bibliography{egbib}
}

\end{document}